\documentclass[letterpaper, 10 pt, conference]{ieeeconf}  

\usepackage{xifthen}
\usepackage{xparse}
\usepackage{subdepth}
\usepackage{leftidx}
\usepackage{bm}

\newcommand{\bbm}{\begin{bmatrix}}
\newcommand{\ebm}{\end{bmatrix}}

\DeclareMathAlphabet{\mbf}{OT1}{ptm}{b}{n}
\newcommand{\mbs}[1]{{\bm{#1}}}

\newcommand{\mbsbar}[1]{{\overline{\boldsymbol{#1}}}}
\newcommand{\mbshat}[1]{{\hat{\boldsymbol{#1}}}}
\newcommand{\mbstilde}[1]{{\tilde{\boldsymbol{#1}}}}

\newcommand{\mbsdot}[1]{{\dot {\boldsymbol{#1}}}}

\newcommand{\mbfbar}[1]{{\overline{\mbf{#1}}}}
\newcommand{\mbfhat}[1]{{\hat{\mbf{#1}}}}
\newcommand{\mbftilde}[1]{{\tilde{\mbf{#1}}}}

\newcommand{\mbfdot}[1]{{\dot{\mbf{#1}}}}

\DeclareMathAlphabet{\mathbfit}{OML}{cmm}{b}{it}
\newcommand{\homo}[1]{{\mathbfit{#1}}}

\newcommand{\mbfh}[1]{{\homo{#1}}}

\newcommand{\vel}[3]{\leftidx{_{#1}}{\mbf v}{\IfValueTF{#2}{_{#2#3\hspace{2pt}}}{}}} 
\newcommand{\veltilde}[3]{\leftidx{_{#1}}{\mbftilde v}{\IfValueTF{#2}{_{#2#3\hspace{2pt}}}{}}} 
\newcommand{\velbar}[3]{\leftidx{_{#1}}{\mbfbar v}{\IfValueTF{#2}{_{#2#3\hspace{2pt}}}{}}} 
\newcommand{\velhat}[3]{\leftidx{_{#1}}{\mbfhat v}{\IfValueTF{#2}{_{#2#3\hspace{2pt}}}{}}} 
\newcommand{\veldot}[3]{\leftidx{_{#1}}{\mbfdot v}{\IfValueTF{#2}{_{#2#3\hspace{2pt}}}{}}} 

\newcommand{\acc}[3]{\leftidx{_{#1}}{\mbf a}{\IfValueTF{#2}{_{#2#3\hspace{2pt}}}{}}} 
\newcommand{\acctilde}[3]{\leftidx{_{#1}}{\mbftilde a}{\IfValueTF{#2}{_{#2#3\hspace{2pt}}}{}}} 
\newcommand{\accbar}[3]{\leftidx{_{#1}}{\mbfbar a}{\IfValueTF{#2}{_{#2#3\hspace{2pt}}}{}}} 

\newcommand{\rotvel}[3]{\leftidx{_{#1}}{\mbs \omega}{\IfValueTF{#2}{_{#2#3\hspace{2pt}}}{}}} 
\newcommand{\rotveltilde}[3]{\leftidx{_{#1}}{\mbstilde \omega}{\IfValueTF{#2}{_{#2#3\hspace{2pt}}}{}}} 
\newcommand{\rotvelbar}[3]{\leftidx{_{#1}}{\mbsbar \omega}{\IfValueTF{#2}{_{#2#3\hspace{2pt}}}{}}} 
\newcommand{\rotvelhat}[3]{\leftidx{_{#1}}{\mbshat \omega}{\IfValueTF{#2}{_{#2#3\hspace{2pt}}}{}}} 
\newcommand{\rotveldot}[3]{\leftidx{_{#1}}{\mbsdot \omega}{\IfValueTF{#2}{_{#2#3\hspace{2pt}}}{}}} 

\newcommand{\T}[2]{\leftidx{}{\mbfh T}{_{#1#2\hspace{2pt}}}} 

\IEEEoverridecommandlockouts                              

\usepackage{graphics} 
\usepackage{epsfig} 
\usepackage{amsmath} 
\usepackage{amssymb}  
\usepackage{amsfonts}
\usepackage[noadjust]{cite}
\usepackage{color}
\usepackage{multirow}
\usepackage{array}
\usepackage{caption}
\usepackage{booktabs}
\usepackage{makecell}
\usepackage{flushend}

\title{\LARGE \bf
DeepFusion: Real-Time Dense 3D Reconstruction for Monocular SLAM using Single-View Depth and Gradient Predictions
}

\author{Tristan Laidlow, Jan Czarnowski and Stefan Leutenegger%
\thanks{The authors are with the Dyson Robotics Laboratory,
        Imperial College London, UK. Corresponding author: Tristan Laidlow, {\tt\small t.laidlow15@imperial.ac.uk}}%
\thanks{Research presented in this paper has been supported by Dyson Technology Ltd.}%
}

\begin{document}

\maketitle
\thispagestyle{empty}
\pagestyle{empty}

\begin{abstract}

While the keypoint-based maps created by sparse monocular simultaneous localisation and mapping (SLAM) systems are useful for camera tracking, dense 3D reconstructions may be desired for many robotic tasks.
Solutions involving depth cameras are limited in range and to indoor spaces, and dense reconstruction systems based on minimising the photometric error between frames are typically poorly constrained and suffer from scale ambiguity.
To address these issues, we propose a 3D reconstruction system that leverages the output of a convolutional neural network (CNN) to produce fully dense depth maps for keyframes that include metric scale.

Our system, DeepFusion, is capable of producing real-time dense reconstructions on a GPU.
It fuses the output of a semi-dense multiview stereo algorithm with the depth and gradient predictions of a CNN in a probabilistic fashion, using learned uncertainties produced by the network.
While the network only needs to be run once per keyframe, we are able to optimise for the depth map with each new frame so as to constantly make use of new geometric constraints.
Based on its performance on synthetic and real-world datasets, we demonstrate that DeepFusion is capable of performing at least as well as other comparable systems.

\end{abstract}

\section{INTRODUCTION}

One of the goals of structure-from-motion (SfM) and visual simultaneous localisation and mapping (SLAM) systems is the incremental creation of 3D scene reconstructions from moving cameras.
Sparse SLAM systems such as MonoSLAM \cite{Davison:etal:PAMI2007}, PTAM \cite{Klein:Murray:ISMAR2009} and ORB-SLAM \cite{Mur-Artal:etal:TRO2017} create sparse 3D maps of keypoints.
While these features are useful for camera tracking, dense reconstructions may be preferred for safe robotic navigation, augmented reality, and manipulation tasks.
Dense reconstruction systems, such as DTAM \cite{Newcombe:etal:ICCV2011} and REMODE \cite{Pizzoli:etal:ICRA2014}, typically create dense scene representations by optimising for the depth values that minimise the photometric error over several frames.
Unfortunately, depth estimation by minimising the photometric error is not well-constrained due the the presence of occlusions, and homogeneous or repeated texture.
To combat this, dense systems often use a regulariser based on planar (\cite{Flint:etal:ICCV2011,Pradeep:etal:ISMAR2013,Concha:etal:RSS2014,Concha:etal:ICRA2014,Concha:Civera:IROS2015}) or smoothness assumptions (\cite{Newcombe:etal:ICCV2011,Pizzoli:etal:ICRA2014}).
LSD-SLAM \cite{Engel:etal:ECCV2014} tackles this issue by reconstructing only those points that have a strong image gradient, but can therefore only produce semi-dense reconstructions.
Monocular reconstruction systems also suffer from an inherent scale ambiguity, as it is not possible to determine a camera's translation from image correspondences alone.
It is possible to address this scale ambiguity by fusing vision-based measurements with readings from an inertial measurement unit (IMU); however, in situations with low accelerations (for example, when the camera is still), scale becomes practically unobservable when using low grade IMUs.
One option to resolve these issues in monocular dense reconstruction is through the use of a depth camera (the approach taken by \cite{Newcombe:etal:ISMAR2011} and \cite{Kerl:etal:IROS2013}), but depth cameras have limited range, consume more power, and do not typically work outdoors or in strong sunlight.

\begin{figure}[t]
  \centering
  \includegraphics[width=1\linewidth,trim={12cm 6cm 10cm 5cm},clip]{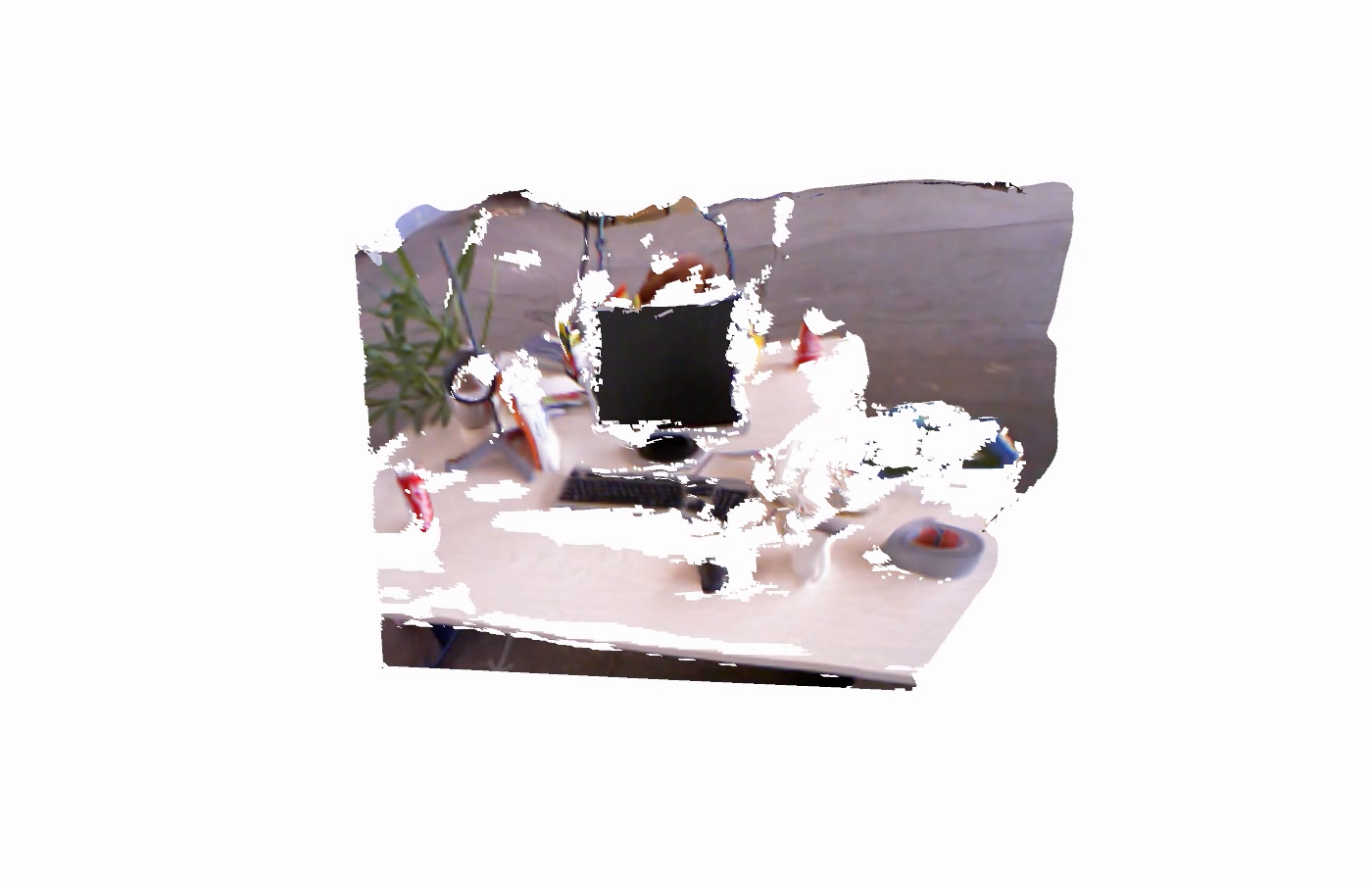}
  \caption{Fusing the depth predictions of a semi-dense multiview stereo system with the depth and gradient predictions of a CNN allows for the creation of fully dense depth maps at scale. The above projected keyframe depth map was created by DeepFusion from only pose estimates and RGB images.}
  \label{fig:teaser}
\end{figure}

\begin{figure*}
    \centering
    \includegraphics[width=\linewidth]{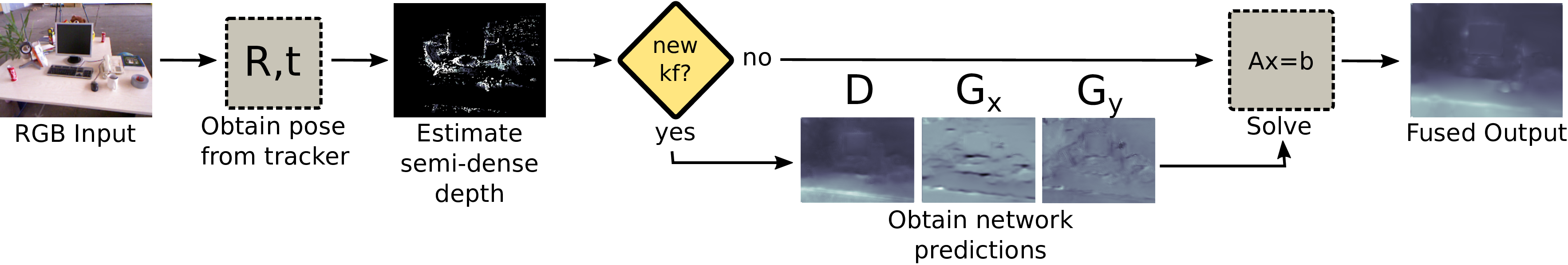}
    \caption{The DeepFusion framework.}
    \label{fig:pipeline}
\end{figure*}

Recently, with the rapid growth of deep learning in computer vision, several data-driven approaches to dense reconstruction have been proposed.
Presumably, the networks used in these systems are able to make predictions about the structure of the scene by leveraging some learned knowledge about observed objects and their spatial relationships.
While many of these approaches advocate a completely end-to-end framework (for example, \cite{Eigen:etal:ICCV2015,Ummenhofer:etal:CVPR2017,Laina:etal:3DV2016,Liu:etal:2015,Li:etal:CVPR2015,Dharmasiri:etal:IROS2017}), there has been some work demonstrating the benefit of combining both geometric constraints and learned priors.
As shown in \cite{Facil:etal:RAL2017}, geometric-based systems perform best on areas of high image gradients (usually on the edges of objects) but struggle with interior areas of low texture, whereas learning-based systems typically do reasonably well on interior points but blur the edges of objects.
Despite the evidence to their complementary nature, however, the best approach to combining learning and geometry remains an open problem.
For example, in \cite{Weerasekera:etal:ICRA2017}, the authors create a dense monocular SLAM system based on DTAM, but use surface normal predictions from a CNN as a strong prior on the depth map rather than use a smoothness regulariser.
Other approaches use geometric reconstructions as an input to a network and then either ``fill in" any missing depth data \cite{Yang:etal:ECCV2018} or refine them \cite{Zhou:etal:ECCV2018}.

One problem with using a network as the final stage in a reconstruction pipeline is that an expensive network pass must be computed every time the underlying geometric information is updated, which may be unacceptably slow for real-time incremental systems.
For this reason, a number of approaches that take network depth predictions and refine them with geometric constraints have been proposed.
In \cite{Facil:etal:RAL2017}, the authors compute a network depth prediction for each keyframe and update a semi-dense multiview stereo depth map with each new frame.
The two depth estimates are then interpolated based on a set of tunable weights related to the image structure.
Another approach \cite{Zhang:Funkhouser:CVPR2018} predicts surface normals and occlusion boundaries for each keyframe image and then attempts to fill in missing values in the output of a depth camera.
Unfortunately, the solve time is too slow to be used on incremental SLAM systems.
In \cite{Yin:etal:ICCV2017}, a CRF is used to refine the regression results.
In \cite{Weerasekera:etal:ICRA2018}, which greatly inspired our work, the authors were able to create accurate dense reconstructions using a fully-connected CRF with network predicted uncertainties from a sparse set of 3D points generated by a monocular SLAM system. Since they use the output of a monocular system to constrain the dense reconstructions, however, the resulting depth maps are ambiguous in scale. In our work, we use the idea of maintaining global consistency by linking neighbouring pixels in our reconstruction through depth gradient predictions and extend it to include the estimation of absolute scale.
Perhaps the work that is most comparable to ours is CNN-SLAM \cite{Tateno:etal:CVPR2017}, which uses a network to predict an at-scale depth map for each keyframe and then refines it through small baseline stereo constraints in real-time. The refinement in CNN-SLAM, however, is done on a per-pixel basis and therefore does not preserve global consistency.

In this paper, we propose DeepFusion, a 3D reconstruction system that is capable of producing dense depth maps at scale in real-time from RGB images and scale-ambiguous poses provided by a monocular SLAM system. We use network predicted depth gradients as a constraint on neighbouring pixels to ensure global consistency in our reconstructions, and learned uncertainties to fuse the different modalities in a probabilistic fashion. Please see Figure \ref{fig:teaser} for an example keyframe reconstruction.

\section{METHOD}

In this section, we describe our system for producing dense reconstructions at scale in real-time. Please see Figure \ref{fig:pipeline} for an overview of the DeepFusion framework.

DeepFusion represents the observed geometry with a series of keyframe depth maps.
With each new RGB image, the system obtains the pose from a monocular SLAM system (ORB-SLAM2 \cite{Mur-Artal:etal:TRO2017} in our implementation) and then updates the semi-dense depth estimates for the active keyframe using the method described in \cite{Engel:etal:ICCV2013}.
If the camera has translated more than $\lambda_{trans}$ or had fewer than $\lambda_{inliers}$ inliers in the semi-dense estimation, a new keyframe is created.

To maintain a high frame rate, our network outputs are only generated once per keyframe.
Using a CNN, we predict the log-depth, log-depth gradients and associated uncertainties from the new keyframe image.
Like \cite{Weerasekera:etal:ICRA2018} and \cite{Eigen:etal:NIPS2014}, we predict log-depths instead of depths or inverse depths because it is numerically better for network prediction (negative values are meaningful) and it has the convenient property that the difference between two log-depths (the gradient of the log-depth image) is the ratio of two depths, which is scale-invariant.
We also choose to predict log-depth gradients in both the x- and y-directions on the image plane rather than surface normals in order to maintain the linearity of the optimisation problem, as this avoids the need for performing dot product and normalisation operations.
Single-view depth prediction is a highly under-constrained problem, and in practice it seems easier for the network to make accurate predictions about fine-grained local geometry rather than absolute per-pixel depth.
For this reason, we predict the absolute log-depth values and the log-depth gradients separately as we want separate uncertainties to reflect the difference in the network's ability at these two different tasks.

If a new keyframe is not created, then the current semi-dense depth map and network outputs are fused to update the current depth map.

\subsection{Network Architecture}

\begin{figure}
    \centering
    \includegraphics[width=\linewidth]{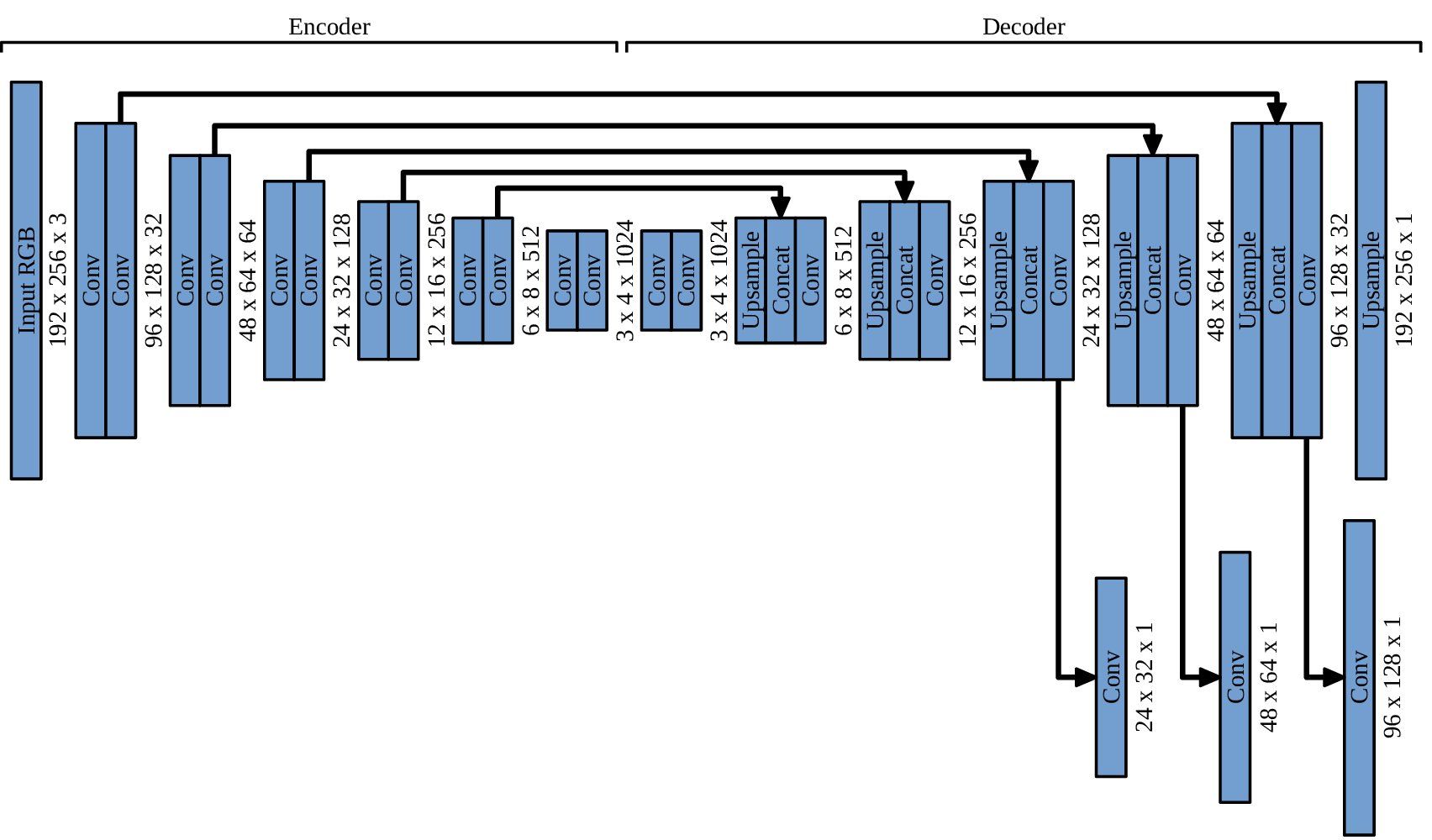}
    \caption[The DeepFusion Network Architecture]{The network consists of a U-Net-style encoder and decoder with five skip connections. During training, outputs are extracted at four different resolutions with losses calculated for each of them. While a single encoder is shared across all tasks, a separate decoder is used for each of the network predictions (depth, depth uncertainty, depth gradients and depth gradient uncertainty).}
    \label{fig:unet}
\end{figure}

For our network, we use a U-Net \cite{Ronneberger:etal:MICCAI2015} architecture with the same dimensions as the one used in \cite{Bloesch:etal:CVPR2018}, except that we add three more identical decoders to predict log-depth uncertainties, log-depth gradients, and log-depth gradient uncertainties in addition to just the log-depths.
All inputs and outputs have a resolution of 256$\times$192.
A diagram of our network architecture is provided in Figure \ref{fig:unet}.

In order to fuse the outputs of the network with the estimates coming from the semi-dense multiview stereo system, we require an uncertainty associated with each pixel in each of the log-depth and gradient images.
To obtain this, we use the method described in \cite{Kendall:Gal:NIPS2017} to have the network learn to predict both a mean and a variance with a maximum likelihood cost function:

\begin{equation}
    \mathcal{L}_{\mathrm{NN}}(\mbs \theta) = \sum_i \frac{(y_i - f_{\mbs \theta,i}(\mbf x))^2}{\sigma_{\mbs \theta,i}(\mbf x)^2} + \log(\sigma_{\mbs \theta,i}(\mbf x)^2),
    \label{eq:loss}
\end{equation}
where $\mbs \theta$ is the set of network weights, $\mbf x$ is the set of input pixels, $y_i$ is the ground truth for pixel $i$, and $f_{\mbs \theta,i}(\mbf x)$ and $\sigma_{\mbs \theta,i}(\mbf x)^2$ are the network's predictions for the mean and variance, respectively.
The total loss is the sum of this loss function for each of the output images.

Like \cite{Bloesch:etal:CVPR2018}, we train our network on the SceneNet RGB-D dataset \cite{McCormac:etal:ICCV2017}, a dataset of 5 million rendered indoor scenes.
We train our network to predict log-depths that have been normalised by the focal length of the SceneNet camera.
In the same manner as CNN-SLAM \cite{Tateno:etal:CVPR2017}, we scale the log-depth predictions of the network by the focal length of the camera used at test time so we can recover absolute scale for images captured by different cameras.
Since all predictions are done in logspace, the gradients (which represent depth ratios) and uncertainties do not need to be scaled.

\subsection{Semi-Dense Estimation}

For our semi-dense multiview stereo component, we implement the depth estimation method from \cite{Engel:etal:ICCV2013}.
For each pixel, $\mbf x_i$, in the keyframe where there is sufficient texture, a search is made along the epipolar line for the depth value, $\mbf d_{\mathrm{semi},i}$ that minimises the sum of squared differences for five equally spaced points.
If there is a current depth estimate for that pixel, the search is conducted over the interval $\mbf d_{\mathrm{semi},i} \pm 2\sigma_{\mathrm{semi},i}$.
Otherwise, the search is conducted over the entire epipolar line.
Given a pixel in the keyframe, $\mbf x_i$, and the estimated poses of the keyframe ($\T{W}{C_{0}}$) and reference frame ($\T{W}{C_{1}}$), the photometric error is given by:
\begin{align}
e_{i} =& \ I_1\left( \mbs \pi \left( \mbf K \, \T{W}{C_{1}}^{-1} \T{W}{C_{0}} \mbs \rho(\mbf x_i, \mbf d_{\mathrm{semi},i}) \right) \right) - I_0(\mbf x_i),
\end{align}
\noindent
where $I_{*}(\cdot)$ is a scalar function that returns the intensity value of a given pixel, $\mbs \pi(\cdot)$ is the projection and dehomogenisation function, $\mbs \rho(\mbf x_i, \mbf d_{\mathrm{semi},i})$ is the back-projection function that returns a homogeneous 3D point for pixel $\mbf x_i$ with a depth $\mbf d_{\mathrm{semi},i}$, and $\mbf K$ is the camera intrinsics matrix.

We approximate the Jacobian of the error function, $\mbf J$, with finite differences:
\begin{equation}
    \mbf J \approx \frac{1}{\Delta \mbf d_{\mathrm{semi},i}}\Delta \mbf e,
\end{equation}
\noindent
where $\mbf e$ is the 5x1 vector containing the photometric error associated with each of the five points. When searching for the minimum value along the epipolar line, even sized steps of 1 pixel length are taken. Once the minimum is found, we interpolate between two steps to find the optimal depth at sub-pixel resolution. The difference in the photometric error at the two endpoints of the interpolation is $\Delta \mbf e$, and the difference between the depths at those points is $\Delta \mbf d_{\mathrm{semi},i}$.

We then approximate the uncertainty of each semi-dense measurement by:
\begin{equation}
    \sigma_i^2 = (\mbf J^T \mbf J)^{-1}.
    \label{eq:semi}
\end{equation}

The semi-dense depth estimates and uncertainties are then converted into logspace to match the network outputs.

\subsection{Optimisation}

To update the current depth prediction, we minimise the following cost function consisting of three terms with each new frame:
\begin{equation}
    c(\mbf d, s) = c_{\mathrm{semi}}(\mbf d, s) + c_{\mathrm{net}}(\mbf d) + c_{\mathrm{grad}}(\mbf d),
\end{equation}
\noindent
where $\mbf d$ is the set of log-depth values to be estimated and $s$ is the scale correction factor.

The semi-dense cost term imposes a unary constraint over the set of pixels where valid semi-dense log-depth values have been estimated:
\begin{subequations}
\begin{align}
    c_{\mathrm{semi}}(\mbf d, s) &= \mbf r_{\mathrm{semi}}^{T}(\mbf d, s) \mbf R_{\mathrm{semi}}^{-1} \mbf r_{\mathrm{semi}}(\mbf d, s), \\
    \mbf r_{\mathrm{semi}, i}(\mbf d_i, s) &= \ln{\mbf d_i} - \ln{s} - \ln{\mbf d_{\mathrm{semi},i}},
\end{align}
\end{subequations}
\noindent
where $\mbf R_{\mathrm{semi},i} = \sigma_i^{2}$, the uncertainty estimated by the approximation given in (\ref{eq:semi}) and $\ln{\mbf d_{\mathrm{semi},i}}$ is the scale-ambiguous log-depth predicted by the semi-dense system for pixel $i$.
Since the semi-dense log-depth estimates are calculated based on the poses provided by a monocular SLAM system, they have an arbitrary scale.
Because the depth map we wish to estimate, $\mbf d$, is to scale, we also need to solve for a scale correction factor, $s$.
With the fully dense per-pixel cost term $c_{\mathrm{net}}$, this scale correction factor becomes observable.

The network depth cost term imposes an additional unary constraint over all pixels of the fused depth map:
\begin{subequations}
\begin{align}
    c_{\mathrm{net}}(\mbf d) &= \mbf r_{\mathrm{net}}^{T}(\mbf d) \mbf R_{\mathrm{net}}^{-1} \mbf r_{\mathrm{net}}(\mbf d), \\
    \mbf r_{\mathrm{net}, i}(\mbf d_i) &= \ln{\mbf d_i} - \ln{\mbf d_{\mathrm{net},i}},
\end{align}
\end{subequations}
\noindent
where $\mbf R_{\mathrm{net},i} = \sigma_{\mbs \theta,\mathrm{depth},i}^2(\mbf x)$, the uncertainty predicted by the network (see (\ref{eq:loss})) and $\ln{\mbf d_{\mathrm{net},i}}$ is the log-depth prediction for pixel $i$ made by the network.
While the network may have a high uncertainty about the absolute depth at any pixel (as the uncertainty is conditioned on the image, it will depend on how close the image is to the training data), this cost term provides a weak prior on the absolute scale of the scene and allows us to estimate the scale correction factor $s$.

In order to maintain global consistency while fusing together the semi-dense and network depth values, we include an additional cost term that imposes pairwise constraints between a given pixel and each of its four neighbours:
\begin{subequations}
\begin{align}
    c_{\mathrm{grad}}(\mbf d) &= \mbf r_{\mathrm{grad},\mathrm{x}}^{T}(\mbf d) \mbf R_{\mathrm{grad},\mathrm{x}}^{-1} \mbf r_{\mathrm{grad},\mathrm{x}}(\mbf d) \\
     &+ \mbf r_{\mathrm{grad},\mathrm{y}}^{T}(\mbf d) \mbf R_{\mathrm{grad},\mathrm{y}}^{-1} \mbf r_{\mathrm{grad},\mathrm{y}}(\mbf d)
    \nonumber \\
    \mbf r_{\mathrm{grad},\mathrm{x},i}(\mbf d_{i+1}, \mbf d_i) &= \ln{\mbf d_{i+1}} - \ln{\mbf d_i} - \mbf g_{\mathrm{x},i} \\
    \mbf r_{\mathrm{grad},\mathrm{y},i}(\mbf d_{i+\mathrm{W}}, \mbf d_i) &= \ln{\mbf d_{i+\mathrm{W}}} - \ln{\mbf d_i} - \mbf g_{\mathrm{y},i}
\end{align}
\end{subequations}
\noindent
where $\mathrm{W}$ is the width of the image, $\mbf g_{\mathrm{x},i}$ and $\mbf g_{\mathrm{y},i}$ are the log-depth gradients in the x- and y-directions predicted by the network, and $\mbf R_{\mathrm{grad,x},i} = \sigma_{\mbs \theta,\mathrm{grad,x},i}^2(\mbf x)$ and $\mbf R_{\mathrm{grad,y},i} = \sigma_{\mbs \theta,\mathrm{grad,y},i}^2(\mbf x)$, the associated predicted uncertainties (see (\ref{eq:loss})).

The semi-dense depth estimates can be very noisy and the network predictions can have extreme outliers which have a significant impact on the final reconstruction.
For this reason, we use the Huber loss function on each of the cost terms.

We solve the system using the Opt \cite{DeVito:etal:ACMTOG2017} optimisation framework. With Opt, we define an energy function for each term in our cost function which are then automatically compiled into GPU optimisation kernels. We compute 10 Gauss-Newton iterations, alternating between solving for the depth and scale. With this setup, we are capable of solving the system for each new frame in real-time.

\section{Experimental Results}

\subsection{Qualitative Results}

\begin{figure*}
  \centering
  \includegraphics[width=1\linewidth]{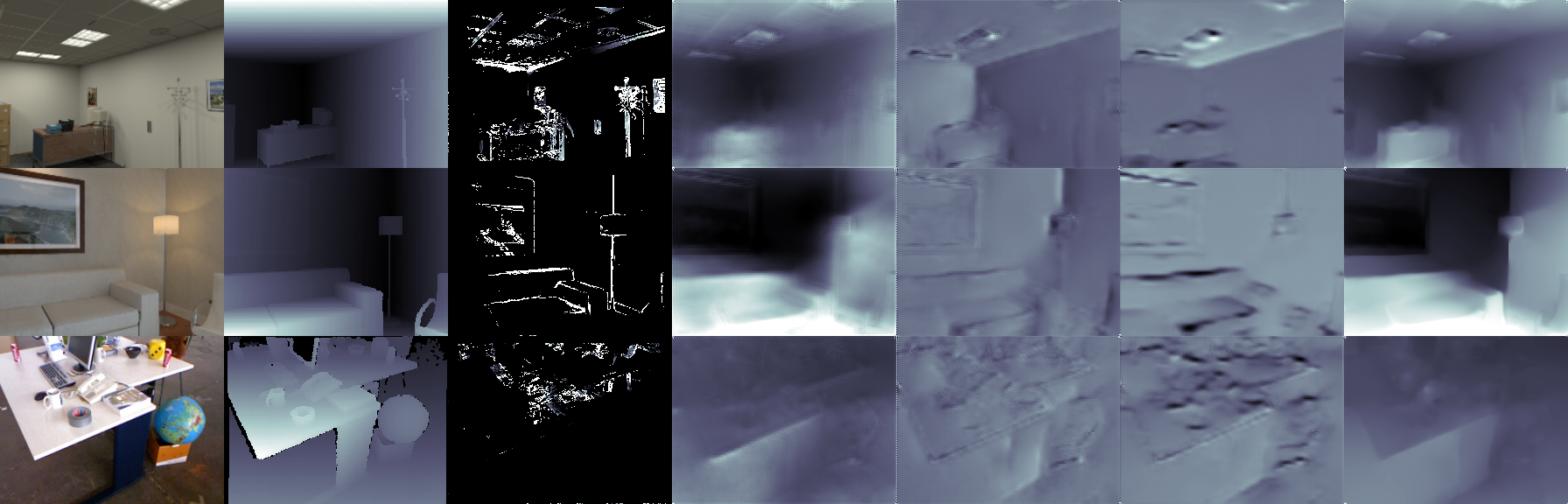}
  \caption{Qualitative results for selected keyframes on the ICL-NUIM Office2 (top), \mbox{ICL-NUIM} LivingRoom1 (middle) and TUM RGB-D fr2\_desk (bottom) sequences. From left to right: input image, ground truth depth, semi-dense depth estimate, network depth prediction, network depth gradient prediction in the x-direction, network depth gradient prediction in the y-direction, and the optimised depth map.}
  \label{fig:examples}
\end{figure*}

Figure \ref{fig:examples} shows some qualitative results from selected keyframes on the ICL-NUIM \cite{Handa:etal:ICRA2014} and TUM RGB-D \cite{Sturm:etal:IROS2012} datasets.
Comparing the network depth predictions with the final fused depth maps shows that including geometric constraints from the semi-dense depth estimation and pairwise pixel constraints from the gradient predictions produces depth maps that are more globally consistent and have fewer blurring artifacts.

\begin{figure*}
  \centering
  \includegraphics[width=1\linewidth]{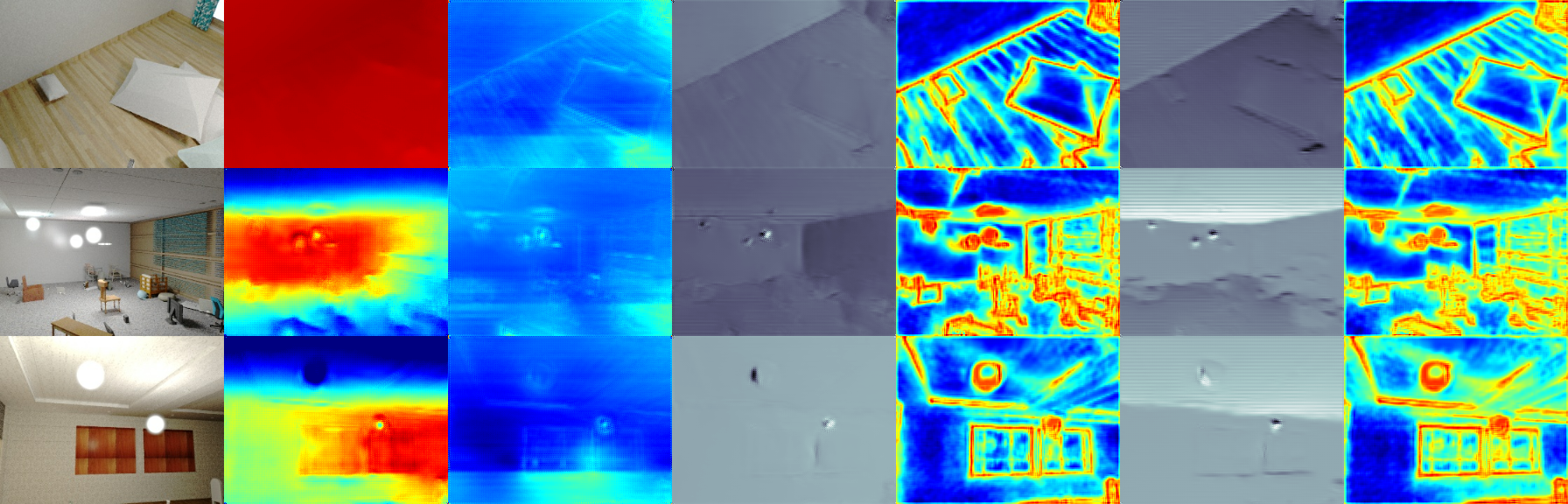}
  \caption{Example network predictions on sample images from the SceneNet RGB-D dataset. From left to right: input image, log-depth prediction, log-depth uncertainty prediction, log-depth gradient prediction in the x-direction, log-depth gradient in the x-direction uncertainty prediction, log-depth gradient prediction in the y-direction, log-depth gradient in the y-direction uncertainty prediction.}
  \label{fig:uncertainties}
\end{figure*}

Figure \ref{fig:uncertainties} shows the network output for sample images in the SceneNet RGB-D dataset \cite{McCormac:etal:ICCV2017}.
The uncertainties associated with the gradient images are clearly largest on areas of high image gradient suggesting that the network has learned that these regions tend to correspond with depth discontinuities or other rapid changes in the depth gradient.

\begin{table*}
  \caption{Comparison of reconstruction accuracy in terms of percentage of correct depth values (within 10\% of ground truth) on ICL-NUIM and TUM RGB-D datasets (TUM/seq1: \textit{fr3/long\_office\_household}, TUM/seq2: \textit{fr3\_nostructure\_texture\_near\_withloop}, TUM/seq3: \textit{fr3/structure\_texture\_far}). LSD-SLAM (BS) is LSD-SLAM bootstrapped with a ground truth depth map, and REMODE uses LSD-SLAM (BS) poses and keyframes.}
  \label{t:recon_compare}
  \begin{center}
      \begin{tabular}{c c c c c c c c}
        \toprule
          \textbf{Sequence} & \textbf{DeepFusion} & \textbf{CNN-SLAM} \cite{Tateno:etal:CVPR2017} & \textbf{LSD-SLAM (BS)} \cite{Engel:etal:ECCV2014} & \textbf{LSD-SLAM} \cite{Engel:etal:ECCV2014} & \textbf{ORB-SLAM} \cite{Mur-Artal:etal:TRO2017} & \textbf{Laina} \cite{Laina:etal:3DV2016} & \textbf{REMODE} \cite{Pizzoli:etal:ICRA2014} \\
        \midrule
          ICL/office0   & \textbf{21.090} &         19.410  & 0.603 & 0.335 & 0.018 &         17.194  &  4.479 \\
          ICL/office1   & \textbf{37.420} &         29.150  & 4.759 & 0.038 & 0.023 &         20.838  &  3.132 \\
          ICL/office2   &         30.180  & \textbf{37.226} & 1.435 & 0.078 & 0.040 &         30.639  & 16.708 \\
          ICL/living0   & \textbf{24.223} &         12.840  & 1.443 & 0.360 & 0.027 &         15.008  &  4.479 \\
          ICL/living1   & \textbf{14.001} &         13.038  & 3.030 & 0.057 & 0.021 &         11.449  &  2.427 \\
          ICL/living2   &         25.235  &         26.560  & 1.807 & 0.167 & 0.014 & \textbf{33.010} &  8.681 \\
          TUM/seq1      &          8.069  &         12.477  & 3.797 & 0.086 & 0.031 & \textbf{12.982} &  9.548 \\
          TUM/seq2      &         14.774  & \textbf{24.077} & 3.966 & 0.882 & 0.059 &         15.412  & 12.651 \\
          TUM/seq3      &         27.200  & \textbf{27.396} & 6.449 & 0.035 & 0.027 &          9.450  &  6.739 \\
        \midrule
          \textbf{Avg.} & \textbf{22.466} &         22.464  & 3.032 & 0.226 & 0.029 &         18.452 &   7.649 \\
        \bottomrule
      \end{tabular}
  \end{center}
  \label{tbl:comparison}
\end{table*}

\subsection{Reconstruction Evaluation}

We evaluate our reconstruction pipeline by comparing it to the results obtained by CNN-SLAM \cite{Tateno:etal:CVPR2017}, a state-of-the-art system that fuses together network predictions and geometric constraints to produce fully dense depth maps.
Following the evaluation procedure of CNN-SLAM, we also compare our results to the depth maps produced by a sparse feature-based monocular system (ORB-SLAM2 \cite{Mur-Artal:etal:TRO2017}), a semi-dense geometric system (LSD-SLAM \cite{Engel:etal:ECCV2014}), a fully dense reconstruction method (REMODE \cite{Pizzoli:etal:ICRA2014}), and a pure deep learning approach (Laina, et al. \cite{Laina:etal:3DV2016}).

As there was no open-source version of CNN-SLAM available at the time of writing, we evaluate DeepFusion on the same sequences used in \cite{Tateno:etal:CVPR2017} and compare with their reported results.
The sequences used for the comparison come from two different datasets: the synthetic ICL-NUIM RGB-D dataset \cite{Handa:etal:ICRA2014} and the real world TUM RGB-D SLAM dataset \cite{Sturm:etal:IROS2012}.
The ICL-NUIM dataset provides rendered depth maps as a ground truth comparison and the TUM RGB-D dataset approximates this with Kinect depth camera images.

As proposed by \cite{Tateno:etal:CVPR2017}, we measure the percentage of estimated depth values that are within 10\% of the corresponding ground truth depth values in order to evaluate both the reconstruction accuracy and density.

The results are presented in Table \ref{tbl:comparison}.
While DeepFusion and CNN-SLAM have approximately the same performance overall, this performance is not evenly distributed over the two datasets.
DeepFusion performs the best on four out of the six ICL-NUIM sequences, whereas CNN-SLAM performs the best on two out of the three TUM RGB-D sequences.
The reason for this most likely has to do with the training data used by the two systems.
Our network is trained on the synthetic SceneNet dataset \cite{McCormac:etal:ICCV2017}, whereas CNN-SLAM uses the network described in \cite{Laina:etal:3DV2016} which is trained on the Kinect-captured NYUv2 \cite{Silberman:etal:ECCV2012}.
In addition, two of the TUM RGB-D sequences used in the comparison consist of the camera moving over flat planes covered in posters, with seemingly little semantic information for the networks to leverage.
These results speak to the importance of keeping the network's training data as close as possible to the domain in which the system will be deployed.
While both DeepFusion and CNN-SLAM clearly outperform the geometry-only systems, the learning-based method proposed by Laina, et al.\ \cite{Laina:etal:3DV2016} also does well, performing the best on two of the nine sequences.
This demonstrates the power of learning-based approaches, but ultimately the systems that can make use of both modalities perform the best overall.

\subsection{Scale Optimisation Evaluation}

While we include the network predicted log-depth values in our optimisation problem to solve for the absolute scale of the reconstruction, there are alternative methods. For instance, in \cite{Weerasekera:etal:ICRA2017}, the authors first predict a scale-ambiguous reconstruction and then scale their reconstruction by finding a least-squares fit with a network predicted depth map. The advantage of our method is that we are able to use the relative uncertainties between the semi-dense estimates and network predictions when performing our fusion.

To demonstrate that our method produces better results, we implemented a version of DeepFusion that optimises for a scale-ambiguous depth map using only the semi-dense and gradient terms of the cost function and then finds a least-squares fit with the network predicted depth and measured its performance on sequences used for our reconstruction evaluation. The results are presented in Table \ref{tbl:ablation}. In seven of the nine sequences, our method outperforms the least-squares post-processing method.

\begin{table}[h]
  \caption{Comparison of scale estimation methods in DeepFusion in terms of percentage of correct depth values (within 10\% of ground truth) on ICL-NUIM and TUM RGB-D datasets (TUM/seq1: \textit{fr3/long\_office\_household}, TUM/seq2: \textit{fr3\_nostructure\_texture\_near\_withloop}, TUM/seq3: \textit{fr3/structure\_texture\_far}). ``Least Squares for Scale Estimation" shows the results when using least squares to align a scale-ambiguous estimation with the network depth prediction to estimate a scaled depth map for each keyframe.}
  \begin{center}
  \begin{tabular}{c c c}
    \toprule
      \textbf{Sequence} & \textbf{DeepFusion} & \makecell{\textbf{Least Squares for} \\ \textbf{Scale Estimation}}\\
    \midrule
      ICL/office0 & \textbf{21.090} &         18.135  \\
      ICL/office1 & \textbf{37.420} &         26.415  \\
      ICL/office2 &         30.180  & \textbf{31.359} \\
      ICL/living0 & \textbf{24.223} &         23.861  \\
      ICL/living1 & \textbf{14.001} &         10.372  \\
      ICL/living2 & \textbf{25.235} &         22.082  \\
      TUM/seq1    &          8.069  & \textbf{ 9.690} \\
      TUM/seq2    & \textbf{14.774} &         14.490  \\
      TUM/seq3    & \textbf{27.200} &         24.047  \\
    \bottomrule
  \end{tabular}
  \end{center}
  \label{tbl:ablation}
\end{table}

\begin{table}[h]
  \caption{Analysis on the importance of pairwise constraints on reconstruction accuracy in terms of percentage of correct depth values (within 10\% of ground truth) on ICL-NUIM and TUM RGB-D datasets (TUM/seq1: \textit{fr3/long\_office\_household}, TUM/seq2: \textit{fr3\_nostructure\_texture\_near\_withloop}, TUM/seq3: \textit{fr3/structure\_texture\_far}). ``No Pairwise Constraints" shows the results when fusing together only the semi-dense and network log-depth values.}
  \begin{center}
  \begin{tabular}{c c c}
    \toprule
      \textbf{Sequence} & \textbf{DeepFusion} & \makecell{\textbf{No Pairwise} \\ \textbf{Constraints}}\\
    \midrule
      ICL/office0 & \textbf{21.090} &         16.641  \\
      ICL/office1 & \textbf{37.420} &         24.633  \\
      ICL/office2 &         30.180  & \textbf{30.899} \\
      ICL/living0 & \textbf{24.223} &         21.643  \\
      ICL/living1 & \textbf{14.001} &         12.774  \\
      ICL/living2 & \textbf{25.235} &         21.772  \\
      TUM/seq1    &          8.069  & \textbf{ 9.469} \\
      TUM/seq2    & \textbf{14.774} &         14.187  \\
      TUM/seq3    & \textbf{27.200} &         23.584  \\
    \bottomrule
  \end{tabular}
  \end{center}
  \label{tbl:nograds}
\end{table}

\subsection{Global Consistency Evaluation}

One of the primary differences between DeepFusion and CNN-SLAM \cite{Tateno:etal:CVPR2017} is that DeepFusion includes a pairwise constraint on neighbouring pixels in order to enforce global consistency on the optimised depth maps.

To show that including these constraints does in fact help produce more accurate reconstructions, we implement a version of DeepFusion that optimises for a depth map with absolute scale using only the network predicted log-depths, the semi-dense log-depth estimates, and their associated uncertainties. We evaluate the performance of this version on the sequences used in the reconstruction evaluation. The results are presented in Table \ref{tbl:nograds}. In seven of the nine sequences, our method outperforms the method that does not enforce global consistency.

\subsection{Timing Evaluation}

To demonstrate the real-time capability of our system, we show the approximate runtimes of each component in Table \ref{tbl:runtime}.
These runtimes were based on our implementation using an Intel Core i7-5820K CPU and a GeForce GTX 980 GPU.

\begin{table}[h]
  \caption{Approximate timing information for key components in the DeepFusion system.}
  \label{t:timing}
  \begin{center}
      \begin{tabular}{c c c c c}
        \toprule
          \textbf{} & \textbf{Semi-Dense} & \textbf{Optimisation} & \textbf{Network Prediction} \\
        \midrule
          Mean & 16ms & 33ms & 45ms \\
          Min  & 1ms & 26ms & 44ms \\
          Max  & 43ms & 47ms & 47ms \\
        \bottomrule
      \end{tabular}
  \end{center}
  \label{tbl:runtime}
\end{table}

\section{Conclusion}

We have presented DeepFusion, a system capable of producing dense 3D reconstructions at scale in real-time.
By formulating a cost function that includes per-pixel losses based on network depth predictions and sparse semi-dense depth estimates with pairwise constraints from network depth gradient predictions we are able to estimate both the shape of the observed scene and its absolute scale.
By predicting both the per-pixel mean and variance, we are able to obtain uncertainties for all network outputs and fuse them together with the geometric constraints in a probabilistic fashion.
Since DeepFusion only requires the network to be run once per keyframe, we are able to maintain real-time capability.

Through a series of experiments on synthetic and real datasets, we demonstrate that DeepFusion performs at least as well as other comparable systems.
Furthermore, through a series of ablation studies, we demonstrate the value of estimating the scale of the depth maps by including the network depth output as a per-pixel constraint in the optimisation and using pairwise constraints to enforce global consistency.

In future work, we will attempt to improve DeepFusion by more thoroughly investigating the impact that certain design choices have on the performance of the algorithm.
For example, as was noted in our comparison with CNN-SLAM, the choice of training data and its similarity to the test environment likely has a significant impact on the quality of the final reconstruction.
Which training dataset to use and how to ensure that the predicted uncertainty reflects differences between training and testing environments are open questions.

Related to the network predictions and their associated uncertainties is the question of how to handle extreme outliers that may be produced by the network. In DeepFusion, this was handled by using a robust cost function (in particular, a Huber loss function) in the optimisation. Whether other cost functions that penalise extreme outliers less severely (such as Tukey) would result in better performance remains to be seen.

Finally, as with all other keyframe-based SLAM systems, there are many trade-offs that need to be considered when choosing keyframe selection criteria. How these trade-offs are impacted when a geometry-based depth estimation is coupled with a learning-based prediction should be investigated.

\bibliographystyle{IEEEtran}
\bibliography{IEEEabrv,robotvision}

\end{document}